# Survival modeling using deep learning, machine learning and statistical methods: A comparative analysis for predicting mortality after hospital admission

Ziwen Wang[1#], Jin Wee Lee[1#], Tanujit Chakraborty[2], Yilin Ning[1], Mingxuan Liu[1], Feng Xie[3], Marcus Eng Hock Ong[4,5], Nan Liu[1,4,6]*

[1] Centre for Quantitative Medicine, Duke-NUS Medical School, Singapore, Singapore

[2] Department of Science and Engineering, Sorbonne University, Abu Dhabi, UAE

[3] Department of Biomedical Data Science, Stanford University, Stanford, USA

[4] Programme in Health Services and Systems Research, Duke-NUS Medical School, Singapore, Singapore

[5] Department of Emergency Medicine, Singapore General Hospital, Singapore, Singapore

[6] Institute of Data Science, National University of Singapore, Singapore, Singapore

[#] Co-first author

*Correspondence: Nan Liu, Centre for Quantitative Medicine, Duke-NUS Medical School, 8 College Road, Singapore 169857, Singapore. Phone: +65 6601 6503. Email: liu.nan@duke-nus.edu.sg

**Keywords:** Survival analysis, Deep learning, Machine learning, Comparative study, All-cause hospital mortality

## Abstract

**Background**

Survival analysis is essential for studying time-to-event outcomes and providing a dynamic understanding of the probability of an event occurring over time. Various survival analysis techniques, from traditional statistical models to state-of-the-art machine learning algorithms, support healthcare intervention and policy decisions. However, there remains ongoing discussion about their comparative performance.

**Methods**

We conducted a comparative study of several survival analysis methods, including Cox proportional hazards (CoxPH), stepwise CoxPH, elastic net penalized Cox model, Random Survival Forests (RSF), Gradient Boosting machine (GBM) learning, AutoScore-Survival, DeepSurv, time-dependent Cox model based on neural network (CoxTime), and DeepHit survival neural network. We applied the concordance index (C-index) for model goodness-of-fit, and integral Brier scores (IBS) for calibration, and considered the model interpretability. As a case study, we performed a retrospective analysis of patients admitted through the emergency department of a tertiary hospital from 2017 to 2019, predicting 90-day all-cause mortality based on patient demographics, clinicopathological features, and historical data.

**Results**

The results of the C-index indicate that deep learning achieved comparable performance, with DeepSurv producing the best discrimination (DeepSurv: 0.893; CoxTime: 0.892; DeepHit: 0.891). The calibration of DeepSurv (IBS: 0.041) performed the best, followed by RSF (IBS: 0.042) and GBM (IBS: 0.0421), all using the full variables. Moreover, AutoScore-Survival, using a minimal variable subset, is easy to interpret, and can achieve good discrimination and calibration (C-index: 0.867; IBS: 0.044).

**Conclusions**

While all models were satisfactory, DeepSurv exhibited the best discrimination and calibration. In addition, AutoScore-Survival offers a more parsimonious model and excellent interpretability.

# 1. Background and Significance

Survival analysis is a statistical field that focuses on studying the time-to-event of interest, such as death, recurrence, or failure[1]. It not only handles censored data effectively but also provides a more dynamic understanding of the probability of an event occurring over time[2]. Binary analysis alone is inadequate for dynamically capturing changes in states, such as qualitative transitions from alive to dead, while classical regression analysis cannot address the complexities associated with censoring[3]. Additionally, event times in medical studies often exhibit heavily skewed distributions, limiting the usefulness of statistical tests that assume a normal data distribution, even when there is no censoring in the dataset[4]. Therefore, the survival outcome differs from other outcomes in that it targets both time-to-event and censored status.

Historically, survival analysis has often relied on statistical models such as the Cox Proportional Hazards (CoxPH) model[5]. More recently, machine learning has made significant progress in the domain of survival analysis. Several promising machine learning algorithms for survival analysis have been developed, such as the random survival forest (RSF)[6] and DeepSurv[7]. Despite the considerable potential of applying machine learning in biomedical research and healthcare, concerns have emerged stemming from the lack of transparency in black-box models[8]. Thus, interpretable machine learning (IML) has emerged as a viable solution and gained prominence as an active area of research[9, 10]. To gain a better understanding of the practical performance of diverse survival models and offer a valuable point of reference for clinical researchers uncertain about the actual performance of complex quantitative models, comparative studies utilizing real-world datasets are essential.

Regarding comparative studies of survival models, there is a vast amount of existing literature[11-14]. These studies have primarily focused on comparisons based on mathematical theory[11] and experimental comparisons using real-world data[12-14]. Currently, most comparative studies based on real medical data are disease-specific and lack comparison with IML models. Additionally, comparative studies using large-scale electronic health records (EHR) data are still primarily concentrated on non-survival models, with limited exploration based on EHR for comparing survival models[15-17]. To fill this gap, we conducted a performance comparison of state-of-the-art algorithms under the survival framework to examine their strengths and weaknesses while also considering model interpretability. We sought to illustrate this by predicting all-cause mortality using real-world healthcare data from the emergency department (ED) of a large tertiary hospital. Secondly, in addition to considering continuous-time machine learning algorithms based on the proportional hazards (PH) assumption, we also explore machine learning methods under non-PH assumption or involve the discretization of time-to-event outcomes. This exploration is vital for determining whether the proportionality assumption is restrictive for commonly available variables when predicting mortality after hospital admission. Third, due to the presence of censoring in survival data, standard evaluation metrics for regression such as mean square error (MSE) and R-squared are not suitable for measuring the performance in survival analysis[2]. Most comparison studies for survival models typically focus on discriminative performance, such as the concordance index (C-index), with few studies evaluating measures of the accuracy of predicted survival functions and calibration, such as the integrated Brier score (IBS). Consequently, our study incorporates survival metrics such as C-index and IBS, which are essential for measuring the model's goodness-of-fit, providing a comprehensive assessment of discrimination, and assisting in the evaluation of calibration and stability of the models.

## 2. Objective

This study aims to compare various survival analysis techniques, ranging from traditional statistical models to state-of-the-art machine learning algorithms. We sought to assess and illustrate the comparative performance of these models by real-world data for survival analysis on 90-day all-cause mortality after hospital admission, providing valuable evidence for researchers and clinicians in choosing appropriate methods.

## 3. Methods

### 3.1 Study Design and Setting

A retrospective cohort study was conducted on patients who visited the ED of Singapore General Hospital (SGH). Singapore, a city-state in Southeast Asia, has a rapidly aging population[18], with about one in every five Singaporeans aged 60 years or older. SGH is a large tertiary hospital in Singapore, which caters to the healthcare needs of this growing demographic. Every year, the SGH ED receives more than 120,000 visits and refers over 36,000 patients for inpatient admissions[19]. EHR data analyzed in this study were obtained from Singapore Health Services. The research received approval from the Centralized Institutional Review Board of Singapore Health Services, and a consent waiver was provided for the collection and analysis of EHR data due to the retrospective nature of the study. Additionally, all data underwent deidentification.

### 3.2 Study population

For this retrospective cohort study, all patients who were hospitalized after visiting the ED of SGH between January 2017 and December 2019 were included. The exclusion criteria were (1) patients younger than 21 years; (2) noncitizen patients who might not have complete medical records in the EHR system; (3) suspected duplication of reports. Available variables included in the study were demographic information, vital signs, lab tests at baseline, comorbidities and history. In the study, the full data set was randomly split into a non-overlapping training cohort (70%), validation cohort (10%, if downstream parameter tuning was needed), and test cohort (20%).

### 3.3 Data processing

The primary outcome of this study was all-cause mortality within 90 days for all hospitalized patients. Individuals that did not die within the designated 90-day period were considered right-censored, including those who were lost to follow-up or did not die during the study period.

Sixty preselected candidate variables were collected based on data availability, clinician opinion, and literature review[20]. The set of variables was classified into categorical and continuous variables since the preprocessing techniques are different for each type. For categorical variables such as race, triage class, malignancy, liver diseases, and diabetes, one-hot encoding was performed to convert the different categories into a binarity feature. As continuous variables were the numerical input features, these were scaled to the standard Gaussian distribution with a mean of 0 and standard deviation of 1 in the deep learning algorithms to support optimizing the model during training and to avoid get stuck in local minima. However, it was found that the normalization of these variables in traditional survival models did not improve the performance or stability in any way. Comorbidities were obtained from hospital diagnoses and discharge records for patients' index emergency admissions that occurred within the five years prior. All diagnoses were recorded using International Classification of Diseases (ICD) codes (ICD-9/ICD-10)[21], a globally adopted diagnostic tool for epidemiological and clinical purposes. The comorbidity variables were extracted using the Charlson Comorbidity Index (CCI)[22], and the algorithms proposed by Quan et al.[23] were applied in this study to link the CCI with the ICD codes. The list of candidate variables and abbreviations were given in the Supplementary **Table S1**.

### 3.4 Statistical analysis

We examined the baseline characteristics of the study population for the training, validation, and test cohorts. In the descriptive summary, counts (percentages) were reported for categorical variables and means (SDs) were reported for continuous variables, where SD stands for standard deviation. In addition, univariate and multivariate CoxPH regression were applied to assess the potential features.

Nine algorithms were selected for training, including three of which are based on the traditional survival methods (CoxPH model[5], the stepwise CoxPH model[24, 25], and the elastic net penalty Cox model [CoxEN][26, 27]), three models from machine learning paradigm (AutoScore-Survival[10], RSF[6] and Gradient Boosting [GBM][28]), and three deep neural network algorithms (DeepSurv[7], time-dependent Cox models [CoxTime][29] and DeepHit[30]). To better understand the assumptions, interpretability, and the necessity for parameter tuning in different methods, these methods are described in **Table 1**.

We fitted models using a training cohort, fine-tuned the parameters based on the validation cohort to achieve optimal performance, and ultimately assessed the performance based on the test cohort. Some traditional models such as CoxPH do not need parameter tuning, therefore the validation set does not play any role in the modelling phase. Among the algorithms considered, a key comparison is the simultaneous conduct of variable selection and estimation, with two traditional methods (stepwise CoxPH and CoxEN) and three machine learning methods (AutoScore-Survival, RSF, and GBM). For the stepwise CoxPH, we consider forward selection based on the Akaike information criterion (AIC). For CoxEN, the tuning parameter $\alpha$ is optimized through five-fold cross-validation using the C-index, which helps obtain optimal model parameter $\alpha$ and identify important features through the EN penalty. As for AutoScore-Survival, the number of variables is determined by the parsimony plot based on the validation cohort, which is established based on variable ranking using RSF. As AutoScore-Survival only provides risk scores, we further estimate mortality probability based on the risk scores. Specifically, the risk score represents the log relative risk, and the Nelson-Aalen estimator then can be used to non-parametrically estimate the baseline hazard function. This ultimately yields the mortality estimator based on AutoScore-Survival. Furthermore, we have not only considered RSF under the full-variable setting but have also conducted variable selection based on variable importance. Similarly, we have performed the same procedure for GBM for comprehensive comparison. For three deep neural network algorithms, the hyperparameters, including learning rate, hidden layers, nodes per layer, dropout, and batch size, were tuned manually to achieve optimal performance, as determined by the training and validation learning rotates. Early stopping regularization was also applied to the deep learning-based algorithms to stop model training when no further improvement in performance on the validation cohort was observed. The final hyperparameters and software used for all techniques are shown in Supplementary **Table S4**.

**3.5 Evaluation Criteria for Survival Prediction**

Specialized performance evaluation metrics for survival analysis were employed, including Harrell's C-index and IBS[31]. Detailed mathematical definitions can be found in Supplementary S-Methods.

**C-index:** The C-index is a standard performance measure in survival analysis, which can be defined as the proportion of concordance pairs in a population[31-33]. It was originally introduced by Frank E. Harrell[32, 33] as a time-independent performance measure, with values ranging from 0 to 1, where a value of 1 indicates perfect performance and 0 represents the worst possible performance. If a model makes random predictions without considering any information from the data, the corresponding C-index would be around 0.5. Typically, a C-index of 0.6 or higher is considered an acceptable level of prediction for most clinical datasets[31, 34].

**IBS:** The Brier score (BS)[35] is frequently utilized to quantify the mean square difference between the observed patient status and the predicted survival probabilities at a particular point in time. To some extent, BS is similar to MSE for regression models, but it is time-dependent due to the dynamic status and employs the inverse probability of censoring weights to deal with censored subjects[36]. For global interpretability, the IBS can be calculated, as it does not consider specific time points but rather takes all available time points as a whole. IBS possesses the appealing feature of simultaneously accounting for discrimination and calibration[37]. It typically ranges from 0 to 1, representing perfect and worst discrimination and calibration, respectively. In practice, a model with IBS below 0.25 is deemed useful[12-14].

# 4. Results

## 4.1 Patient characteristics

Sixty variables, routinely available for patients admitted to the hospital from ED, were included in this study, consisting of 43 continuous variables and 17 categorical variables (see Supplementary **Table S1**). A total of 124,873 inpatient admission episodes were finally included, with 87,412 episodes in the training cohort, 12,487 episodes in the validation cohort, and 24,974 in the testing cohort. Patients in these three cohorts had similar characteristics and outcome distributions (Supplementary **Table S2**). The Kaplan-Meier curve for the overall population was plotted in **Figure 1**. Among the included episodes, 112,118 (89.8%) survived for more than 90 days when censored at the end of the designated 90-day observation period. In contrast, 12,755 (10.2%) episodes had a death event within 90 days, with a median time to death of 28 days (IQR: 11 - 53) and a mean time to death of 34 days (SD: 25.7). Supplementary **Table S3** summarizes the candidate variables under different event statuses. Continuous variables were presented as either mean and standard deviation or median and interquartile range. Student t-tests were applied to test continuous variables that followed normal distributions, while the Mann–Whitney U test was used for non-normal continuous variables. Categoric variables were expressed as counts and percentages, and tested using either chi-square or Fisher exact tests. There were statistically significant differences in the features between diffident event status (**Table S3**).

## 4.2 Prediction and interpretability

We report the results of all the prediction algorithms discussed in Section 3.4 by predicting the 90-day all-cause mortality of inpatients in the ED dataset. **Table 2** presents the results of the univariate and multivariate analyses of the Cox model for all variables. From univariate analysis, it is observed that basophils absolute count, the total blood cell count, rheumatic, paralysis, and diabetes without chronic complications did not exhibit statistical significance at the 0.001 level. All variables other than these can be considered risk factors for inpatients. As a result, selecting a parsimonious model based only on *p*-value is challenging.

For the traditional methods regarding variable selection, as shown in Supplementary **Figure S2**, the optimal parameter $\alpha$ for CoxEN is 0.0074, resulting in the retention of 26 variables. Supplementary **Figure S3** presents the estimated shrunken coefficients for all the retained variables. Another traditional method, the stepwise CoxPH, retained 50 out of the total 60 variables. Supplementary **Figure S4** shows the 50 coefficients of all retained variables in the stepwise CoxPH model, ranked by log relative hazards.

For machine learning-based methods, AutoScore-Survival selected 16 variables based on the balance between model performance measured by integrated the area under the receiver operating characteristic curve and the complexity represented by the number of variables, as shown in the parsimony plot (Supplementary **Figure S1**). The sixteen-variable survival score as tabulated in Supplementary **Table S5**, taking into account malignancy, total cell count, age, respiratory rate, diastolic BP, albumin, SAO$_2$, heart rate, troponin T quantitative, blood urea nitrogen, systolic BP, sodium, chloride, basophils absolute count and red cell distribution width. For RSF, the variable importance is consistent with that obtained in step one of AutoScore-Survival, shown in Supplementary **Figure S5**. Additionally, Supplementary **Figure S6** presents the variable importance for the GBM method.

For the deep neural network algorithms (DeepSurv, CoxTime, and DeepHit), all variables were used for training, and the losses of partial log-likelihood are visualized in Supplementary **Figure S7 – S9** respectively. It can be observed that the errors at each iteration gradually decrease until stabilize in both the training and validation cohorts.

### 4.3 Performance comparisons

To compare the performance of all the prediction algorithms on the ED dataset, we report the mean and 95% confidence interval (CI) of the C-index and IBS for all algorithms using the same testing cohort that was completely isolated from the training cohort.

**Table 3** shows the performance results of the C-index. Compared with the standard CoxPH model (C-index: 0.879), three deep neural network algorithms showed better discrimination of all-cause mortality in hospitalized patients (C-index of DeepSurv: 0.893; CoxTime: 0.892; DeepHit: 0.891), with DeepSurv having the highest C-index of 0.890. However, the above three deep neural network algorithms used full candidate variables and lacked interpretability. The AutoScore-Survival (C-index: 0.867) provides an interpretable score and achieves discrimination comparable to CoxPH regression while utilizing a minimal number of variables. Although RSF achieved a high C-index of 0.889 with the same variables as AutoScore-Survival, it is a black box comprised of an ensemble of several survival trees, and the provided variable importance lacks the same level of interpretability as a scoring table. Similar to RSF, GBM exhibits high discrimination (C-index: 0.891). Although it provides variable importance, understanding the specific effect of each variable on outcome is also quite intricate.

The IBS in **Table 3** showed that the consistency between the model prediction and the actual observation for all-cause mortality within 90 days was best for the DeepSurv model (IBS: 0.0414; 95% CI: 0.0403 – 0.0427), followed by the RSF (IBS: 0.0426; 95%: 0.0415 – 0.0437) and CoxTime (IBS: 0.0437; 95% CI: 0.0426 – 0.0449). It is interesting to note that AutoScore-Survival (IBS: 0.0456, 95% CI: 0.0443 – 0.0468), which reduces the number of variables from 60 to only 16, provides better calibration than CoxEN (using 26 variables), GBM (using 16 variables), and even outperforms DeepHit (using the full set of 60 variables).

**4.4 Model visualization**

To assess the statistical significance of different models' performance, we further conducted the Multiple Comparisons with the Best (MCB) test[38] for C-index and IBS measures. The MCB plot was developed to realize the visualization of the model, which added great clinical application value in choosing an appropriate model. As shown in **Figure 2,** Figures A and B correspond to models using all variables, while Figures C and D correspond to models with variable selection. We simulate 20 observations for each method with specific mean and confidence intervals. This non-parametric test calculates average ranks and critical distances for performance measures. In **Figure 2**, for models using all variables, the MCB results reveal that DeepSurv performs best in terms of C-index and IBS measures. For models using variable selection, it is observed that GBM exhibits the best discriminative ability, while RSF demonstrates robust calibration. The critical region of the best-performing model (shaded region) represents the reference value for the test. Models with critical regions overlapping the reference value do not demonstrate statistically significant performance differences, while models with the critical region above this value significantly underperform compared to the best-performing one.

## 5. Discussion

As advancements in data collection techniques continue to generate larger clinical datasets, it is crucial to identify the best methods for analyzing complex survival data. In this study, we provide a comprehensive benchmark evaluation of deep learning, several machine learning, and traditional statistical models for survival analysis on 90 days of all-cause mortality after hospital admission using data from ED admissions at SGH. The results of our research demonstrate that traditional statistical methods tend to have better interpretability, while machine learning and deep learning algorithms have superior discrimination. However, deep learning algorithms such as DeepHit have challenges for calibration. Second, in the models using all variables, DeepSurv performed best in both discrimination and calibration. In the models using variable selection, GBM exhibits the best discrimination, while RSF demonstrates robust calibration. However, both of them have moderate interpretability. In addition, among the compared methods, the AutoScore-Survival is the most easily interpretable parsimonious model and has competitive calibration performance. Therefore, medical researchers and clinicians can choose suitable models based on varying requirements.

Traditional survival models can infer the effects of variables on survival time, providing a source of interpretability for survival prediction. Nevertheless, addressing the complexity of real-world data brings challenges for traditional survival models, such as infeasible fits or reduced predictive accuracy due to overfitting[39]. On the other hand, deep learning and machine learning can produce more accurate survival predictions by considering complex, non-linear relationships between large amounts of information such as disease statuses and feature profiles. For instance, the CoxPH model assumes a PH structure, which makes modeling feature interactions inappropriate. Some machine learning algorithms are not constrained by this assumption and strong collinearity between variables do not affect their prediction accuracy. As a result, traditional statistical methods tend to have better interpretability, while machine learning methods have superior predictive performance.

Despite the favorable predictive performance demonstrated by black-box machine learning models, their lack of interpretability is a significant limitation when it comes to clinical applications as physicians may find it difficult to understand how and why the model arrives at specific predictions. In contrast, a score derived from AutoScore-Survival provides clearer clinical interpretability while exhibiting competitive predictive performance. Since the importance of each scoring variable is transparent, clinicians are more easily able to understand and trust the model outputs, contributing to the effectiveness and acceptability of such interpretable methods in real-world decision making in medical and health applications.

Another crucial component of model development and comparison is variable selection, which is particularly important when dealing with clinical data due to the sheer number of patient variables at hand. Obtaining the most useful and parsimonious set of variables improves model coherence and contributes to a better understanding of the primary risk factors associated with all-cause mortality for inpatients. We compared three different approaches for variable selection based on the CoxPH model. One common approach is the stepwise method, which calculates the AIC or BIC by adding and/or eliminating variables to find the optimal model. A popular alternative to the stepwise method is the elastic net penalty method. Unlike the stepwise approach, the elastic net penalty method includes all variables in the model but shrinks some regression coefficients to zero. The third approach is based on RSF variable importance ranking, which is the first step of the AutoScore-Survival estimation process. We observed that all three methods conclude that Malignancy is the most important variable and has a significant impact on survival time. AutoScore selected the minimum number of variables for interpretation and also demonstrated a comparative advantage in terms of calibration accuracy.

This study has several limitations. First, as it is a single-site study, the performance of variable methods may exhibit variability in different healthcare settings. Therefore, the results might not be directly generalizable to other populations. Further investigation can be conducted to explore models trained on benchmark data and generalize them to other clinical datasets. Second, the benchmark dataset used in this study is based on large-scale EHR data, including routinely collected variables. However, certain features were not taken into consideration, for example, the Glasgow Coma Scale (GCS) score is excluded due to a high missing rate of 78%. GCS is considered an important predictive factor in an ED setting and could potentially improve the performance of these survival models. Despite these limitations, this study provides valuable evidence for researchers in choosing the most appropriate method.

## 6. Conclusion

This study sought to compare the performance of state-of-the-art survival methods for predicting all-cause mortality using real-world healthcare data. Overall, all methods demonstrated satisfactory performance. Traditional statistical methods tend to have better interpretability, ensemble machine learning algorithms provide good discrimination and calibration, and deep learning algorithms have a superior discriminative ability but calibration can be challenging. In addition, interpretable machine learning like AutoScore-Survival showed comparable performance, providing a more parsimonious model and superior interpretability.

**Data availability statement**

All data created or analyzed during this investigation are included in this article and its supplementary information files.

**Authorship contribution statement**

NL conceived the idea and supervised the study. ZW and JW performed the methods and analyzed the results. ZW drafted the manuscript. All authors interpreted the data, critically revised the manuscript, and approved the final manuscript.

**Declaration of Competing Interest**

The authors declare that they have no known competing financial interests or personal relationships that could have appeared to influence the work reported in this paper.

**Table 1.** Description of various methods

| Classification | Models | Proportional hazards Assumption | Interpretability | Parameter tuning |
|---|---|---|---|---|
| Traditional statistical method | CoxPH model | Yes | High | No |
| | Stepwise CoxPH | Yes | High | No |
| | CoxEN | Yes | High | No |
| Ensemble machine learning | RSF | No | Moderate | Yes |
| | GBM | No | Moderate | Yes |
| Interpretability machine learning | AutoScore-Survival | Yes | High | Yes |
| Feedforward deep neural network | DeepSurv | Yes | Low | Yes |
| | CoxTime | No | Low | Yes |
| | DeepHit | No | Low | Yes |

**Table 2**: Univariate and multivariate analysis of the Cox model for survival probability.

| Variables | Univariate Cox Regression | | Multivariate Cox Regression | |
|---|---|---|---|---|
| | Hazard Ratio (95%) | P-value | Hazard Ratio (95%) | P-value |
| Age (years) | 1.034 (1.032 – 1.035) | < 0.001 | 1.033 (1.031 – 1.035) | <0.001 |
| Gender (Male) | 1.202 (1.153 – 1.253) | < 0.001 | 1.151 (1.102 – 1.202) | <0.001 |
| Race | | | | |
|   Chinese | - | - | - | - |
|   Indian | 0.578 (0.532 – 0.629) | <0.001 | 0.990 (0.908 – 1.079) | 0.8140 |
|   Malay | 0.747 (0.697 – 0.802) | <0.001 | 1.260 (1.172 – 1.356) | <0.001 |
|   Others | 0.539 (0.471 – 0.617) | <0.001 | 0.804 (0.701 – 0.921) | 0.0016 |
| Triage class | | | | |
|   P1 | - | - | - | - |
|   P2 | 0.509 (0.487 – 0.531) | <0.001 | 0.763 (0.728 – 0.801) | <0.001 |
|   P3&P4 | 0.174 (0.157 – 0.194) | <0.001 | 0.468 (0.419 – 0.524) | <0.001 |
| Diastolic BP | 0.974 (0.972 – 0.975) | <0.001 | 1.006 (1.004 – 1.008) | <0.001 |
| Systolic BP | 0.983 (0.983 – 0.984) | <0.001 | 0.993 (0.992 – 0.994) | <0.001 |
| FIO2 | 3.759 (3.148 – 4.488) | <0.001 | 1.661 (1.383 – 1.994) | <0.001 |
| Heart rate | 1.020 (1.019 – 1.021) | <0.001 | 1.008 (1.006 – 1.009) | <0.001 |
| Respiration rate | 1.119 (1.112 – 1.125) | <0.001 | 1.044 (1.036 – 1.052) | <0.001 |
| SAO2 | 0.973 (0.971 – 0.975) | <0.001 | 0.986 (0.984 – 0.989) | <0.001 |
| Temperature | 0.937 (0.913 – 0.962) | <0.001 | 0.843 (0.822 – 0.866) | <0.001 |
| Blood albumin | 0.868 (0.865 – 0.871) | <0.001 | 0.952 (0.948 – 0.956) | <0.001 |
| Basophils absolute count | 0.981 (0.866 – 1.112) | 0.7685 | 1.018 (0.913 – 1.136) | 0.7443 |
| Basophils cell | 0.360 (0.329 – 0.394) | <0.001 | 0.834 (0.769 – 0.904) | <0.001 |
| Bicarbonate | 0.953 (0.948 – 0.958) | <0.001 | 0.979 (0.972 – 0.985) | <0.001 |
| Chloride | 0.929 (0.926 – 0.932) | <0.001 | 0.935 (0.929 – 0.941) | <0.001 |
| Serum creatinine | 1.000 (1.000 – 1.001) | <0.001 | 0.999 (0.999 – 0.999) | <0.001 |
| Eosinophils absolute count | 0.787 (0.716 – 0.865) | <0.001 | 1.049 (1.014 – 1.084) | 0.0055 |
| Eosinophils cell | 0.897 (0.886 – 0.907) | <0.001 | 0.970 (0.959 – 0.981) | <0.001 |
| Blood glucose | 1.009 (1.005 – 1.013) | <0.001 | 0.982 (0.977 – 0.987) | <0.001 |
| Hematocrit | 0.918 (0.915 – 0.920) | <0.001 | 1.252 (1.193 – 1.314) | <0.001 |
| Hemoglobin | 0.783 (0.777 – 0.790) | <0.001 | 0.564 (0.497 – 0.641) | <0.001 |
| Lymphocytes absolute | 0.624 (0.605 – 0.643) | <0.001 | 1.014 (1.005 – 1.023) | <0.001 |
| Lymphocytes cell | 0.935 (0.933 – 0.938) | <0.001 | 0.956 (0.951 – 0.959) | 0.0030 |
| MCHB | 1.023 (1.016 – 1.030) | <0.001 | 0.776 (0.684 – 0.881) | <0.001 |
| MCHC | 0.872 (0.862 – 0.884) | <0.001 | 1.440 (1.282 – 1.617) | <0.001 |
| Mean corpuscular volume | 1.028 (1.025 – 1.031) | <0.001 | 1.087 (1.042 – 1.133) | <0.001 |
| Mean platelet volume | 1.057 (1.035 – 1.080) | <0.001 | 1.051 (1.028 – 1.075) | <0.001 |
| Monocytes absolute | 1.036 (1.030 – 1.043) | <0.001 | 1.008 (0.991 – 1.026) | 0.3494 |
| Monocytes cell | 0.980 (0.974 – 0.987) | <0.001 | 0.975 (0.967 – 0.982) | <0.001 |
| Neutrophils absolute count | 1.026 (1.025 – 1.027) | <0.001 | 1.003 (0.998 – 1.008) | 0.2744 |
| Neutrophils cell | 1.042 (1.040 – 1.044) | <0.001 | 0.978 (0.974 – 0.982) | <0.001 |
| Platelet count | 1.000 (1.000 – 1.001) | <0.001 | 1.000 (0.999 – 1.000) | 0.0281 |
| Serum potassium | 1.328 (1.289 – 1.368) | <0.001 | 1.098 (1.063 – 1.135) | <0.001 |
| Red blood cell | 0.476 (0.465 – 0.488) | <0.001 | 0.632 (0.517 – 0.773) | <0.001 |
| Red cell distribution width | 1.214 (1.208 – 1.220) | <0.001 | 1.107 (1.098 – 1.117) | <0.001 |
| Serum sodium | 0.937 (0.934 – 0.940) | <0.001 | 1.031 (1.023 – 1.038) | <0.001 |
| Total absolute count | 1.008 (1.008 – 1.009) | <0.001 | 1.009 (0.972 – 1.047) | 0.6382 |
| Total blood cells count | 1.373 (1.085 – 1.738) | 0.0083 | 1.280 (1.019 – 1.608) | 0.0339 |
| Troponin T | 1.000 (1.000 – 1.000) | <0.001 | 1.000 (1.000 – 1.000) | 0.0514 |
| Blood urea nitrogen | 1.045 (1.043 – 1.047) | <0.001 | 1.029 (1.026 – 1.032) | <0.001 |
| White blood cell | 1.008 (1.008 – 1.009) | <0.001 | 0.987 (0.952 – 1.024) | 0.4982 |
| MI | 2.894 (2.714 – 3.086) | <0.001 | 1.664 (1.551 – 1.785) | <0.001 |
| CHF | 1.766 (1.655 – 1.855) | <0.001 | 1.023 (0.954 – 1.098) | 0.5210 |
| PVD | 1.728 (1.591 – 1.877) | <0.001 | 1.533 (1.405 – 1.672) | <0.001 |
| Stroke | 1.125 (1.052 – 1.202) | <0.001 | 1.409 (1.309– 1.517) | <0.001 |
| Dementia | 1.765 (1.612 – 1.933) | <0.001 | 1.301 (1.183 – 1.431) | <0.001 |
| Pulmonary | 1.213 (1.127 – 1.306) | <0.001 | 1.068 (0.988 – 1.154) | 0.0999 |
| Rheumatic | 0.751 (0.605 – 0.932) | 0.0094 | 0.989 (0.796 – 1.230) | 0.9226 |
| PUD | 1.619 (1.435 – 1.827) | <0.001 | 0.685 (0.605 – 0.776) | <0.001 |
| Paralysis | 1.180 (1.066 – 1.306) | 0.0014 | 1.144 (1.025 – 1.278) | 0.0169 |
| Renal | 1.545 (1.477 – 1.616) | <0.001 | 1.074 (1.011 – 1.141) | 0.0215 |
| Malignancy | | | | |
|    None | - | | | |
|    Local tumor leukemia and lymphoma | 3.319 (3.112 – 3.539) | <0.001 | 2.205 (2.061 – 2.360) | <0.001 |
|    Metastatic solid tumor | 10.526 (10.064 – 11.009) | <0.001 | 6.689 (6.343 – 7.055) | <0.001 |
| LiverD | | | | |
|    None | - | | | |
|    Mild | 1.535 (1.393 – 1.691) | <0.001 | 1.255 (1.136 – 1.387) | <0.001 |
|    Severe | 2.955 (2.655 – 3.289) | <0.001 | 1.372 (1.222 – 1.540) | <0.001 |
| Diabetes | | | | |
| None | - | | | |
| Diabetes without chronic complications | 0.895 (0.782 – 1.024) | 0.1062 | 0.981 (0.855 – 1.126) | 0.7852 |
| Diabetes with complications | 1.262 (1.209 – 1.317) | <0.001 | 0.989 (0.940 – 1.041) | 0.6820 |
| ED# | 1.032 (1.028 – 1.036) | <0.001 | 0.977 (0.959 – 0.997) | 0.0212 |
| INP# | 1.081 (1.075 – 1.086) | <0.001 | 1.053 (1.028 – 1.079) | <0.001 |
| SURG# | 1.180 (1.165 – 1.196) | <0.001 | 1.012 (0.991 – 1.033) | 0.2698 |
| HD# | 1.244 (1.205 – 1.285) | <0.001 | 0.902 (0.868 – 0.939) | <0.001 |
| ICU# | 1.421 (1.342 – 1.505) | <0.001 | 1.084 (1.011 – 1.162) | 0.0227 |

**Table 3**: Performance of different methods with/without variable selection mechanisms.

| **Methods** | **No. of Variables** | **Evaluation Criteria** | |
|---|---|---|---|
| | | **C-index** | **CI (95%)** |
| **CoxPH** | 60 | 0.879 (0.0031) | 0.873 – 0.885 |
| **CoxEN** | 26 | 0.875 (0.0035) | 0.868 – 0.882 |
| **Stepwise CoxPH** | 50 | 0.879 (0.0033) | 0.872 – 0.886 |
| **AutoScore-Survival** | 16 | 0.867 (0.0031) | 0.861 – 0.873 |
| **RSF** | 16 | 0.876 (0.0032) | 0.871 – 0.882 |
| **RSF** | 60 | 0.889 (0.0028) | 0.883 – 0.895 |
| **GBM** | 16 | 0.880 (0.0028) | 0.874 – 0.885 |
| **GBM** | 60 | 0.891 (0.0034) | 0.884 – 0.898 |
| **DeepSurv** | 60 | 0.893 (0.0032) | 0.886 – 0.899 |
| **CoxTime** | 60 | 0.891 (0.0027) | 0.886 – 0.896 |
| **DeepHit** | 60 | 0.892 (0.0031) | 0.886 – 0.898 |
| | **No. of Variables** | **IBS** | **CI (95%)** |
| **CoxPH** | 60 | 0.0428 (0.0008) | 0.0414 – 0.0443 |
| **CoxEN** | 26 | 0.0445 (0.0010) | 0.0426 – 0.0467 |
| **Stepwise CoxPH** | 50 | 0.0436 (0.0009) | 0.0416 – 0.0457 |
| **AutoScore-Survival** | 16 | 0.0439 (0.0008) | 0.0425 – 0.0456 |
| **RSF** | 16 | 0.0425 (0.0008) | 0.0411 – 0.0440 |
| **RSF** | 60 | 0.0418 (0.0008) | 0.0405 – 0.0434 |
| **GBM** | 16 | 0.0445 (0.0008) | 0.0427 – 0.0459 |
| **GBM** | 60 | 0.0421 (0.0010) | 0.0406 – 0.0442 |
| **DeepSurv** | 60 | 0.0406 (0.0009) | 0.0390 – 0.0423 |
| **CoxTime** | 60 | 0.0429 (0.0008) | 0.0412 – 0.0443 |
| **DeepHit** | 60 | 0.0489 (0.0010) | 0.0470 – 0.0511 |

**Figure 1**. Kaplan-Meier curve of training and testing cohorts. There was no statistically significant difference between the survival of training and testing cohorts in the log-rank test (p = 0:14).

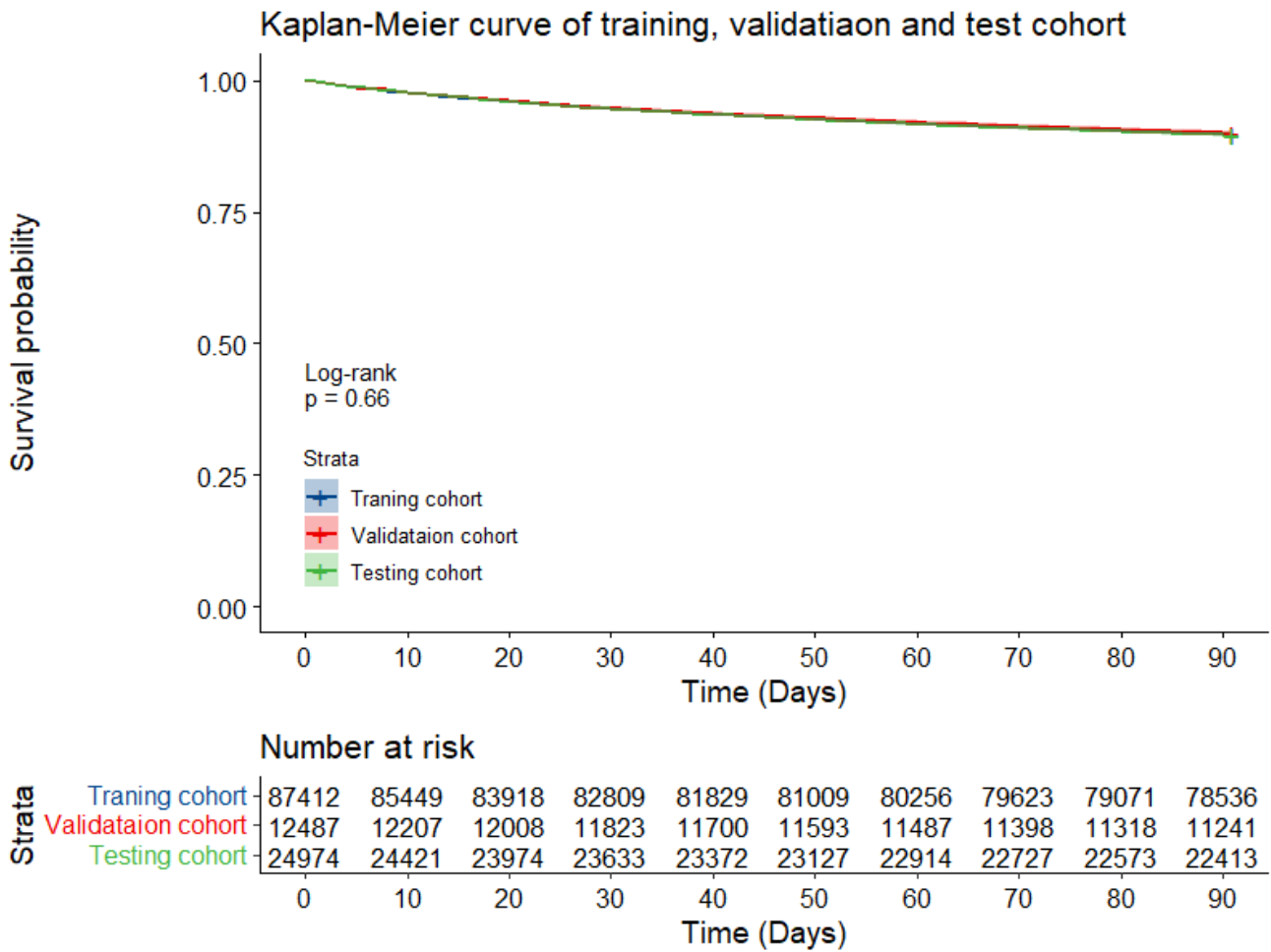

**Figure 2**. Visualization of the MCB test in terms of C-index (left) and IBS (right) measures. Figures A and B represent models using full variables, while Figures C and D correspond to models with variable selection. The dot into the middle is the mean rank of the method, such as GBM (16)-1.45 indicates the mean rank as 1.45 for GBM, which selected 16 variables, and the line above and below the dot is the critical distance. The red dots indicate that the performance difference is insignificant and the black dots represent a significant difference.

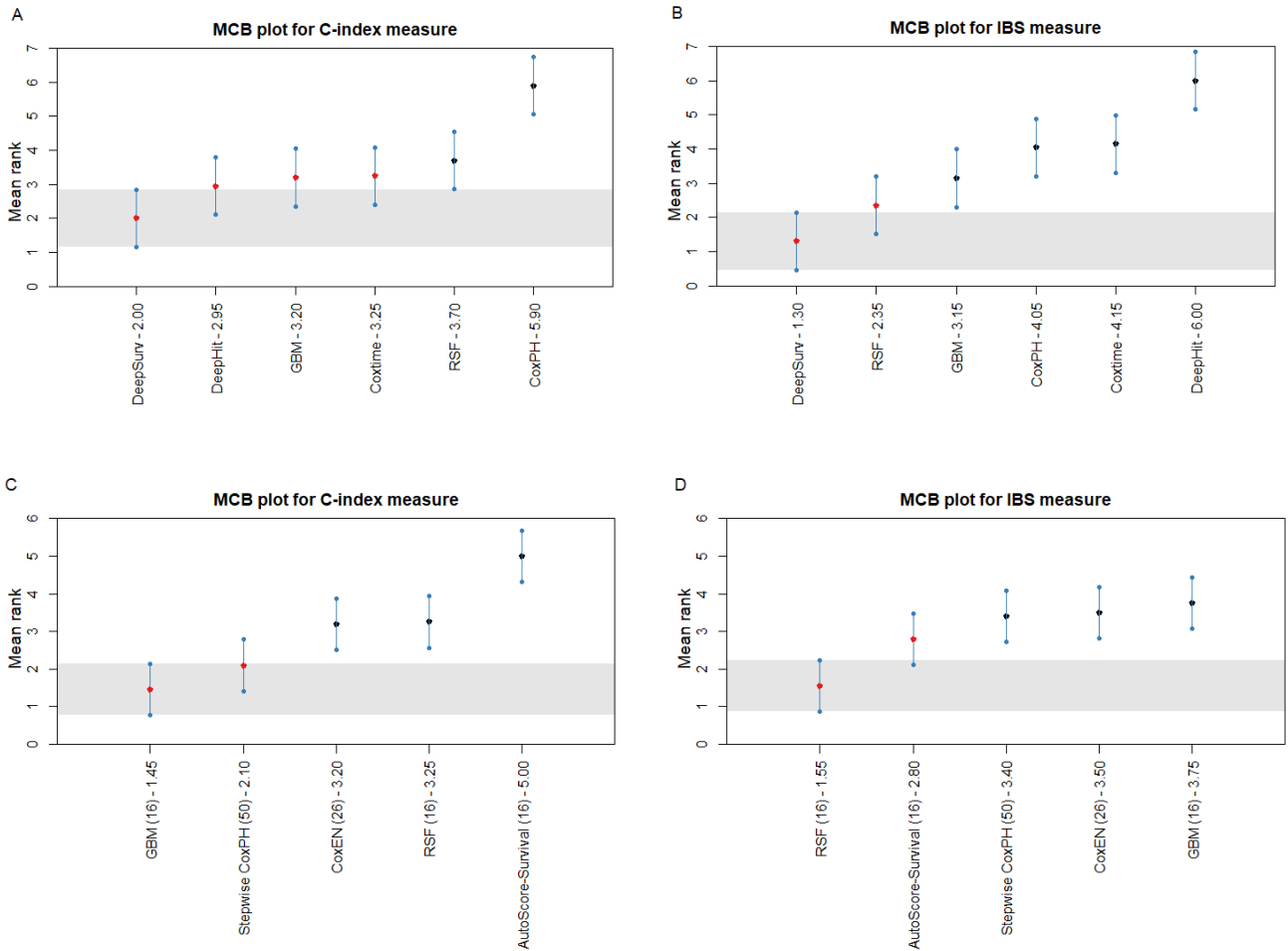

# Supplementary Material

Table S1. List of candidate variables and their abbreviations.

Table S2. Characteristics of study participants for the training, validation, and testing cohorts.

Table S3. Characteristics of study participants under different survival statues.

Table S4. Description of various benchmark methods.

Table S5. Sixteen-variable score for all-cause mortality for the inpatient dataset.

Figure S1. Parsimony plot on the validation cohort based on AutoScore-Survival.

Figure S2. Choosing penalty strength $\alpha$ by concordance index (based on C-index).

Figure S3. Variable importance on the validation cohort based on the CoxEN model.

Figure S4. Variable importance based on the stepwise CoxPH model.

Figure S5. Variable importance based on RSF.

Figure S6. Variable importance based on GBM.

Figure S7. A plot about the loss of partial log-likelihood on training and validation cohort for DeepSurv algorithm.

Figure S8. A plot about the loss of partial log-likelihood on training and validation cohort for CoxTime algorithm.

Figure S9. A plot about the loss of partial log-likelihood on training and validation cohort for DeepHit algorithm.

S-Method. Description of the benchmarking accuracy measures.

**Table S1.** List of candidate variables and their abbreviations.

| Classification | Variables (Abbreviation) | Categorical/ Continuous |
|---|---|---|
| Demographics Information | Age | Continuous |
| | Gender | Categorical |
| | Race | Categorical |
| Vital signs | Triage class | Categorical |
| | Diastolic blood pressure (Diastolic BP) | Continuous |
| | Systolic blood pressure (Systolic BP) | Continuous |
| | Fraction of inspiration oxygen (FIO2) | Categorical |
| | Heart rate | Continuous |
| | Respiratory rate | Continuous |
| | Arterial oxygen saturation (SAO2) | Continuous |
| | Temperature | Continuous |
| Laboratory results | Blood albumin (ALB) | Continuous |
| | Basophils absolute count (BAS#) | Continuous |
| | Basophils cell (BAS%) | Continuous |
| | Bicarbonate (HCO3-) | Continuous |
| | Chloride (Cl-) | Continuous |
| | Serum creatinine (Cr) | Continuous |
| | Eosinophils absolute count (EOS#) | Continuous |
| | Eosinophils cell (EOS%) | Continuous |
| | Blood glucose (GLU) | Continuous |
| | Hematocrit (HCT) | Continuous |
| | Hemoglobin (HGB) | Continuous |
| | Lymphocytes absolute (LYMPH#) | Continuous |
| | Lymphocytes cell (LYMPH%) | Continuous |
| | Mean corpuscular hemoglobin (MCHB) | Continuous |
| | Mean corpuscular hemoglobin concentration (MCHC) | Continuous |
| | Mean corpuscular volume (MCV) | Continuous |
| | Mean platelet volume (MPV) | Continuous |
| | Monocytes absolute count (MONO#) | Continuous |
| | Monocytes cell (MONO%) | Continuous |
| | Neutrophils absolute count (NEUT#) | Continuous |
| | Neutrophils cell (NEUT%) | Continuous |
| | Platelet count (PLT) | Continuous |
| | Potassium (K+) | Continuous |
| | Red blood cell (RBC) | Continuous |
| | Red cell distribution width (RDW) | Continuous |
| | Serum sodium (Na+) | Continuous |
| | Total absolute count (TAC) | Continuous |
| | Total blood cells count (TCC) | Continuous |
| | Troponin T quantitative (Troponin T) | Continuous |
| | Blood urea nitrogen (BUN) | Continuous |
| | White blood cell (WBC) | Continuous |
| Comorbidities | Myocardial infarction (MI) | Categorical |
| | Congestive heart failure (CHF) | Categorical |
| | Peripheral vascular diseases (PVD) | Categorical |
| | Stroke | Categorical |
| | Dementia | Categorical |
| | Chronic pulmonary diseases (PulmonaryD) | Categorical |

|  | Rheumatic diseases (RheumaticD) | Categorical |
|  | Peptic ulcer disease (PUD) | Categorical |
|  | Hemiplegia or paraplegia (Paralysis) | Categorical |
|  | Renal diseases (Renal) | Categorical |
|  | Malignancy | Categorical |
|  | Liver diseases (LiverD) | Categorical |
|  | Diabetes | Categorical |
| **History information** | Emergency admissions in the past year (ED#) | Continuous |
|  | Inpatient admission in the past year (INP#) | Continuous |
|  | Surgeries in the past year (SURG#) | Continuous |
|  | HUD admission in the past year (HD#) | Continuous |
|  | ICU admission in the past year (ICU#) | Continuous |

**Table S2.** Characteristics of study participants for the training, validation, and testing cohorts.

| Characteristic | Over all | Training cohort | Validation cohort | Test cohort |
|---|---|---|---|---|
| **No. of participants** | 124,873 | 87,412 | 12,487 | 24,974 |
| **Age (years)** | 65.38 (16.64) | 65.34 (16.62) | 65.54 (16.69) | 65.42 (16.69) |
| **Gender** | | | | |
|   Male | 61,845 (49.5%) | 43,223 (49.4%) | 6,250 (50.1%) | 12,372 (49.5%) |
|   Female | 63,018 (50.5%) | 44,189 (51.6%) | 6,237 (49.9%) | 12,602 (50.5%) |
| **Race** | | | | |
|   Chinese | 92,360 (74.0%) | 64,656 (74.0%) | 9,251 (74.1%) | 18,453 (73.9%) |
|   Indian | 12,856 (10.3%) | 8,975 (10.3%) | 1,335 (10.7%) | 2,546 (10.2%) |
|   Malay | 14,668 (11.7%) | 10,266 (11.7%) | 1,398 (11.2%) | 3,004 (12.0%) |
|   Others | 4,989 (4.0%) | 3,515 (4.0%) | 503 (4.0%) | 971 (3.9%) |
| **Triage class** | | | | |
|   P1 | 28,630 (22.9%) | 20,005 (22.9%) | 2,887 (23.1%) | 5,738 (23.0%) |
|   P2 | 80,248 (64.3%) | 56,165 (64.3%) | 8,052 (64.5%) | 16,031 (64.2%) |
|   P3 & P4 | 15,995 (12.8%) | 11,242 (12.9%) | 1,548 (12.4%) | 3,205 (12.8%) |
| **Diastolic blood pressure (mmHg)** | 72.54 (14.38) | 72.51 (14.30) | 72.68 (14.49) | 72.60 (14.59) |
| **Systolic blood pressure (mmHg)** | 136.86 (27.50) | 136.76 (27.40) | 137.34 (27.48) | 136.97 (27.86) |
| **FIO2** | | | | |
|   =21 | 124,315 (99.6%) | 87,021 (99.6%) | 12,431 (99.6%) | 24,863 (99.6%) |
|   >21 | 558 (0.4%) | 391 (0.4%) | 56 (0.4%) | 111 (0.4%) |
| **Pulse (bpm)** | 85.47 (18.33) | 85.50 (18.33) | 85.35 (18.48) | 85.42 (18.26) |
| **Respiratory rate (cpm)** | 18.15 (2.11) | 18.15 (2.11) | 18.13 (2.13) | 18.16 (2.11) |
| **SAO2 (%)** | 97.42 (4.14) | 97.42 (4.15) | 97.40 (4.34) | 97.43 (4.00) |
| **Temperature (°C)** | 36.71 (0.81) | 36.71 (0.82) | 36.70 (0.81) | 36.70 (0.82) |
| **Blood albumin (g/L)** | 38.31 (3.80) | 38.31 (3.79) | 38.30 (3.81) | 38.28 (3.84) |
| **Basophils absolute ($10^9$/L)** | 0.05 (0.25) | 0.04 (0.20) | 0.04 (0.11) | 0.05 (0.40) |
| **Basophil cell (%)** | 0.47 (0.33) | 0.47 (0.33) | 0.47 (0.34) | 0.47 (0.33) |
| **Bicarbonate (mmol/L)** | 23.18 (3.60) | 23.16 (3.60) | 23.22 (3.57) | 23.21 (3.62) |
| **Chloride (mmol/L)** | 101.72 (5.27) | 101.71 (5.25) | 101.81 (5.25) | 101.68 (5.34) |
| **Serum creatinine ($\mu$mol/L)** | 144.12 (188.33) | 143.98 (188.35) | 143.33 (189.38) | 145.02 (187.75) |
| **Eosinophils absolute ($10^9$/L)** | 0.18 (0.39) | 0.17 (0.40) | 0.17 (0.28) | 0.18 (0.41) |
| **Eosinophil cell (%)** | 2.00 (2.75) | 1.99 (2.74) | 2.02 (2.74) | 2.00 (2.77) |
| **Blood glucose (mmol/L)** | 8.27 (4.73) | 8.28 (4.77) | 8.21 (4.47) | 8.26 (4.74) |
| **Hematocrit (%)** | 36.66 (6.52) | 36.68 (6.53) | 36.63 (6.51) | 36.61 (6.50) |
| **Hemoglobin (g/dL)** | 12.16 (2.32) | 12.17 (2.32) | 12.15 (2.31) | 12.14 (2.31) |
| **Lymph absolute ($10^9$/L)** | 1.65 (2.30) | 1.64 (1.63) | 1.65 (1.52) | 1.68 (4.00) |
| **Lymph cell (%)** | 18.80 (10.37) | 18.79 (10.36) | 18.82 (10.34) | 18.83 (10.41) |
| **MCHB (pg/g)** | 29.16 (3.08) | 29.15 (3.08) | 29.20 (3.04) | 29.17 (3.07) |
| **MCHC (g/L)** | 33.10 (1.45) | 33.09 (1.45) | 33.10 (1.42) | 33.10 (1.44) |
| **Mean corpuscular volume (fL)** | 87.96 (7.85) | 87.94 (7.85) | 88.08 (7.77) | 87.99 (7.86) |
| **Mean platelet volume (fL)** | 9.94 (0.96) | 9.94 (0.96) | 9.94 (0.96) | 9.93 (0.97) |
| **Monocytes absolute ($10^9$/L)** | 0.73 (0.86) | 0.73 (0.93) | 0.73 (0.70) | 0.72 (0.65) |
| **Monocytes cell (%)** | 7.75 (3.29) | 7.76 (3.29) | 7.76 (3.40) | 7.72 (3.21) |
| **Neutrophils absolute count ($10^9$/L)** | 7.22 (4.89) | 7.23 (4.84) | 7.19 (4.70) | 7.21 (5.17) |
| **Neutrophils cell (%)** | 70.59 (12.80) | 70.60 (12.82) | 70.53 (12.73) | 70.59 (12.79) |
| **Platelet count ($10^9$/L)** | 263.28 (105.05) | 263.52 (105.26) | 262.00 (102.66) | 263.10 (105.52) |
| **Serum potassium (mmol/L)** | 4.12 (0.62) | 4.13 (0.62) | 4.12 (0.62) | 4.12 (0.62) |
| **Red blood cell ($10^{12}$/L)** | 4.20 (0.83) | 4.21 (0.83) | 4.19 (0.83) | 4.20 (0.82) |
| **Red cell distribution width (%)** | 14.28 (2.23) | 14.28 (2.23) | 14.26 (2.19) | 14.28 (2.24) |
| **Serum sodium (mmol/L)** | 136.27 (5.00) | 136.26 (4.98) | 136.38 (5.00) | 136.27 (5.06) |
| **Total absolute count ($10^9$/L)** | 9.95 (7.69) | 9.94 (7.49) | 9.89 (5.95) | 9.97 (9.00) |
| **Total blood cells count ($10^9$/L)** | 100.00 (0.29) | 100.00 (0.34) | 100.00 (0.2) | 100.00 (0.01) |
| **Troponin T quantitative (mmol/L)** | 18.78 (156.20) | 18.98 (164.51) | 16.92 (75.84) | 18.99 (156.16) |
| **Blood urea nitrogen (mmol/L)** | 8.16 (7.00) | 8.16 (7.00) | 8.09 (6.97) | 8.18 (7.03) |
| **White blood cell ($10^9$/L)** | 9.93 (7.70) | 9.94 (7.51) | 9.87 (5.97) | 9.95 (9.02) |
| **MI** | 6,082 (4.9%) | 4,258 (4.9%) | 556 (4.5%) | 1,268 (5.1%) |
| **CHF** | 8,864 (7.1%) | 6,201 (7.1%) | 882 (7.1%) | 1,781 (7.1%) |
| **PVD** | 5,082 (4.1%) | 3,635 (4.2%) | 452 (3.6%) | 995 (4.0%) |
| **Stroke** | 12,472 (10.0%) | 8,671 (9.9%) | 1,310 (10.5%) | 2,491 (10.0%) |
| **Dementia** | 4,205 (3.4%) | 2,902 (3.3%) | 439 (3.5%) | 864 (3.5%) |
| **Pulmonary** | 9,089 (7.3%) | 6,379 (7.3%) | 903 (7.2%) | 1,807 (7.2%) |
| **Rheumatic** | 1,517 (1.2%) | 1,061 (1.2%) | 146 (1.2%) | 310 (1.2%) |
| **PUD** | 2,433 (1.9%) | 1,708 (2.0%) | 249 (2.0%) | 476 (1.9%) |
| **Paralysis** | 4,705 (3.8%) | 3,269 (3.7%) | 496 (4.0%) | 940 (3.8%) |
| **Renal** | 28,340 (22.7%) | 19,845 (22.7%) | 2,765 (22.1%) | 5,730 (22.9%) |
| **AllCancer** | | | | |
| None | 104,495 (83.7%) | 73,132 (83.7%) | 10,484 (84.0%) | 20,879 (83.6%) |
| Local tumor leukemia and lymphoma | 9,562 (7.7%) | 6,717 (7.7%) | 933 (7.5%) | 1,912 (7.7%) |
| Metastatic solid tumor | 10,816 (8.7%) | 7,563 (8.7%) | 1,070 (8.6%) | 2,183 (8.7%) |
| **Liver disease** | | | | |
| None | 118,771 (95.1%) | 8,3147 (95.1%) | 11,888 (95.2%) | 23,736 (95.0%) |
| Mild | 4,207 (3.4%) | 2,930 (3.4%) | 415 (3.3%) | 862 (3.5%) |
| Severe | 1,985 (1.5%) | 1,335 (1.5%) | 184 (1.5%) | 376 (1.5%) |
| **Diabetes** | | | | |
| None | 79,940 (64.0%) | 55,947 (64.0%) | 8,047 (64.4%) | 15,946 (63.9%) |
| Diabetes without chronic complications | 3,770 (3.0%) | 2,571 (2.9%) | 382 (3.1%) | 817 (3.3%) |
| Diabetes with complications | 41,163 (33.0%) | 28,894 (33.1%) | 4,058 (32.5%) | 8,211 (32.9%) |
| **Number of emergency admissions** | 1.56 (3.25) | 1.55 (3.21) | 1.57 (3.30) | 1.59 (3.36) |
| **Number of inpatient admission** | 1.01 (2.11) | 1.01 (2.10) | 1.01 (2.10) | 1.02 (2.13) |
| **Number of surgeries** | 0.28 (0.95) | 0.28 (0.95) | 0.28 (0.91) | 0.29 (0.95) |
| **Number of HUD admission** | 0.08 (0.44) | 0.08 (0.43) | 0.09 (0.47) | 0.08 (0.45) |

| **Number of ICU admission** | 0.02 (0.21) | 0.02 (0.21) | 0.02 (0.22) | 0.02 (0.22) |
|---|---|---|---|---|
| **Survival times** | 85.14 (19.22) | 85.14 (19.23) | 85.25 (19.05) | 85.10 (19.26) |
| **status** | | | | |
|     True | 12,755 (10.2%) | 8,932 (10.2%) | 1,249 (10.0%) | 2,574 (10.3%) |
|     False | 112,118 (89.8%) | 78,480 (89.8%) | 11,238 (90.0%) | 22,400 (89.7%) |

*Continuous variables are presented as Mean (SD); binary/categorical variables are presented as Count (%)

**Table S3.** Characteristics of study participants under different survival statues.

| | Status | | |
|---|---|---|---|
| **Characteristic** | **False** | **Ture** | *P*-value |
| **No. of participants** | 112,118 (89.8%) | 12,755 (10.2%) | |
| **Age (years)** | 67.00 [56.00, 78.00] | 74.00 [64.00, 83.00] | <0.001 |
| **Gender** | | | <0.001 |
|   Male | 54,887 (49.0%) | 6,958 (54.6%) | |
|   Female | 57,231 (51.0%) | 5,797 (45.4%) | |
| **Race** | | | <0.001 |
|   Chinese | 82,003 (73.1%) | 10,357 (81.2%) | |
|   Indian | 12,001 (10.7%) | 845 (6.6%) | |
|   Malay | 13,436 (12.0%) | 1,232 (9.7%) | |
|   Others | 4,668 (4.2%) | 321 (2.5%) | |
| **Triage class** | | | <0.001 |
|   P1 | 23,766 (21.2%) | 4,864 (38.1%) | |
|   P2 | 72,822 (65.0%) | 7,371 (57.8%) | |
|   P3 & P4 | 15,475 (13.8%) | 520 (4.1%) | |
| **Diastolic blood pressure (mmHg)** | 72.00 [63.00, 81.00] | 67.00 [58.00, 76.00] | <0.001 |
| **Systolic blood pressure (mmHg)** | 135.00 [119.00, 154.00] | 124.00 [107.00, 141.00] | <0.001 |
| **FIO2** | | | <0.001 |
|   >21 | 111,731 (99.7%) | 12,584 (98.7%) | |
|   =21 | 387 (0.3%) | 171 (1.3%) | |
| **Pulse (bpm)** | 83.00 [72.00, 96.00] | 91.00 [78.00, 105.00] | <0.001 |
| **Respiratory rate (cpm)** | 18.00 [17.00, 19.00] | 18.00 [18.00, 20.00] | <0.001 |
| **SAO2 (%)** | 98.00 [97.00, 99.00] | 98.00 [96.00, 99.00] | <0.001 |
| **Temperature (°C)** | 36.60 [36.10, 37.10] | 36.50 [36.10, 37.10] | <0.001 |
| **Blood albumin (g/L)** | 39.00 [39.00, 39.00] | 39.00 [31.00, 39.00] | <0.001 |
| **Basophils absolute ($10^9$/L)** | 0.04 [0.03, 0.05] | 0.04 [0.02, 0.05] | <0.001 |
| **Basophil cell (%)** | 0.40 [0.30, 0.60] | 0.40 [0.20, 0.50] | <0.001 |
| **Bicarbonate (mmol/L)** | 23.40 [21.40, 25.30] | 23.10 [19.90, 25.40] | <0.001 |
| **Chloride (mmol/L)** | 103.00 [100.00, 105.00] | 100.00 [95.00, 103.00] | <0.001 |
| **Serum creatinine ($\mu$mol/L)** | 77.00 [63.00, 114.00] | 88.00 [63.00, 168.00] | <0.001 |
| **Eosinophils absolute ($10^9$/L)** | 0.11 [0.04, 0.21] | 0.06 [0.01, 0.14] | <0.001 |
| **Eosinophil cell (%)** | 1.40 [0.50, 2.60] | 0.70 [0.10, 1.60] | <0.001 |
| **Blood glucose (mmol/L)** | 6.50 [5.90, 8.80] | 6.80 [6.10, 9.10] | <0.001 |
| **Hematocrit (%)** | 38.10 [33.50, 41.00] | 33.00 [27.90, 38.30] | <0.001 |
| **Hemoglobin (g/dL)** | 12.70 [11.00, 13.70] | 10.80 [9.10, 12.60] | <0.001 |
| **Lymph absolute ($10^9$/L)** | 1.59 [1.06, 2.05] | 1.10 [0.70, 1.65] | <0.001 |
| **Lymph cell (%)** | 19.20 [11.60, 25.20] | 11.50 [6.60, 19.20] | <0.001 |
| **MCHB (pg/g)** | 29.80 [28.20, 30.80] | 29.80 [27.80, 31.10] | <0.001 |
| **MCHC (g/L)** | 33.30 [32.40, 34.00] | 32.90 [31.80, 33.70] | <0.001 |
| **Mean corpuscular volume (fL)** | 88.40 [84.90, 92.00] | 89.10 [85.20, 94.30] | <0.001 |
| **Mean platelet volume (fL)** | 9.90 [9.40, 10.40] | 9.90 [9.30, 10.60] | **0.030** |
| **Monocytes absolute ($10^9$/L)** | 0.62 [0.50, 0.84] | 0.67 [0.50, 0.96] | <0.001 |
| **Monocytes cell (%)** | 7.30 [6.00, 9.10] | 7.30 [5.30, 9.20] | <0.001 |
| **Neutrophils absolute count ($10^9$/L)** | 5.70 [4.41, 8.47] | 7.29 [5.26, 11.06] | <0.001 |
| **Neutrophils cell (%)** | 69.20 [62.30, 79.10] | 78.10 [68.40, 85.90] | <0.001 |
| **Platelet count ($10^9$/L)** | 248.00 [204.00, 308.00] | 248.00 [175.00, 335.00] | <0.001 |
| **Serum potassium (mmol/L)** | 4.00 [3.80, 4.40] | 4.10 [3.80, 4.60] | <0.001 |
| **Red blood cell ($10^{12}$/L)** | 4.36 [3.78, 4.73] | 3.70 [3.11, 4.34] | <0.001 |
| **Red cell distribution width (%)** | 13.40 [12.90, 14.60] | 15.20 [13.70, 17.40] | <0.001 |
| **Serum sodium (mmol/L)** | 137.00 [135.00, 139.00] | 135.00 [131.00, 138.00] | <0.001 |
| **Total absolute count ($10^9$/L)** | 8.58 [6.93, 11.30] | 9.55 [7.34, 13.56] | <0.001 |
| **Total blood cells count ($10^9$/L)** | 100.00 [100.00, 100.00] | 100.00 [100.00, 100.00] | **0.032** |
| **Troponin T quantitative (ng/mL)** | 0.14 [0.14, 13.00] | 0.14 [0.14, 27.00] | <0.001 |
| **Blood urea nitrogen (mmol/L)** | 5.50 [4.30, 8.60] | 7.90 [4.90, 14.00] | <0.001 |
| **White blood cell ($10^9$/L)** | 8.58 [6.91, 11.30] | 9.55 [7.29, 13.56] | <0.001 |
| **MI** | 4,600 (4.1%) | 1,482 (11.6%) | <0.001 |
| **CHF** | 7,367 (6.6%) | 1,497 (11.7%) | <0.001 |
| **PVD** | 42,41 (3.8%) | 841 (6.6%) | <0.001 |
| **Stroke** | 11,064 (9.9%) | 1,408 (11.0%) | <0.001 |
| **Dementia** | 3,473 (3.1%) | 732 (5.7%) | <0.001 |
| **Pulmonary** | 7,998 (7.1%) | 1,091 (8.6%) | <0.001 |
| **Rheumatic** | 1,394 (1.2%) | 123 (1.0%) | 0.007 |
| **PUD** | 2,044 (1.8%) | 389 (3.0%) | <0.001 |
| **Paralysis** | 4,160 (3.7%) | 545 (4.3%) | 0.002 |
| **Renal** | 24,469 (21.8%) | 3,871 (30.3%) | <0.001 |
| **AllCancer** | | | <0.001 |
|   None | 98,530 (87.9%) | 5,965 (46.8%) | |
|   Local tumor leukemia and lymphoma | 7,838 (7.0%) | 1,724 (13.5%) | |
|   Metastatic solid tumor | 5,750 (5.1%) | 5,066 (39.7%) | |
| **Liver disease** | | | <0.001 |
|   None | 107,104 (95.5%) | 11,667 (91.5%) | |
|   Mild | 3,614 (3.2%) | 593 (4.6%) | |
|   Severe | 1,400 (1.2%) | 495 (3.9%) | |
| **Diabetes** | | | <0.001 |
|   None | 72,322 (64.5%) | 7,618 (59.7%) | |
|   Diabetes without chronic complications | 3,439 (3.2%) | 331 (2.6%) | |
|   Diabetes with complications | 36,357 (32.4%) | 4,806 (37.7%) | |

*Continuous variables are presented as Median [IQR]; binary/categorical variables are presented as Count (%).

**Table S4.** Description of various benchmark methods

| Models | Description | Hyperparameters | Software (Package) |
|---|---|---|---|
| Cox proportional hazards model (CoxPH) | Traditional statistical method | None | R (survival) |
| Stepwise CoxPH | Traditional statistical method | None | Python (lifelines) |
| Cox model with elastic net penalty (CoxEN) | Traditional statistical method | Penalty = 'Elastic net', Alpha = '0.05' | Python (scikit-surv) |
| AutoScore-Survival | Interpretability machine learning | Number of trees = 500, Number of variables tuned through performance-based parsimony plot | R (AutoScore) |
| Random survival forest (RSF) | Ensemble machine learning | Number of trees = 500 | R (randomForestSRC) |
| Gradient Boosting (GBM) | Ensemble machine learning | Number of trees = 500 | Python (scikit-surv) |
| DeepSurv | Feedforward deep neural network | Activation = "relu", Drop out = 0.3, Learning rate = "0.001", Batch size = 256, Epoch = 64, Loss = partial log-likelihood, Early stopping = True, Optimizer = "Adam", Notes = [128, 64, 32] | Python (Pycox) |
| CoxTime | Feedforward deep neural network | Activation = "relu", Drop out = 0.2, Learning rate = "0.001", Batch size = 512, Epoch = 64, Loss = partial log-likelihood, Early stopping = True, Optimizer = "Adam" Notes = [128, 64, 32] | Python (Pycox) |
| DeepHit | Feedforward deep neural network | Activation = "relu", Drop out = 0.2, Learning rate = "0.001", Batch size = 512, Epoch = 64, Loss = partial log-likelihood, Early stopping = True, Optimizer = "Adam" Notes = [128, 64, 32] | Python (Pycox) |

**Table S5.** Sixteen-variable score for all-cause mortality for the inpatient dataset.

| Variables | Interval | Point |
|---|---|---|
| Malignancy | NA | 0 |
| | 1local | 7 |
| | 2metastatic | 15 |
| Total cell count (TCC) | < 100 | 0 |
| | ≥ 100 | 6 |
| Age | [21, 41) | 0 |
| | [41, 58) | 7 |
| | [58, 76) | 10 |
| | [76, 85) | 13 |
| | ≥ 85 | 17 |
| Respiratory rate | <16 | 2 |
| | [16, 17) | 0 |
| | [17, 18) | 1 |
| | [18, 20) | 1 |
| | ≥ 20 | 3 |
| Diastolic BP | < 79 | 0 |
| | [79, 91) | 1 |
| | ≥ 91 | 2 |
| Blood albumin (ALB) | < 34 | 10 |
| | [34, 39) | 5 |
| | [39, 41) | 6 |
| | ≥ 41 | 0 |
| SAO2 | < 95 | 4 |
| | [95, 97) | 1 |
| | ≥ 97 | 0 |
| Heart rate | < 75 | 0 |
| | [75, 94) | 1 |
| | [94, 109) | 2 |
| | ≥ 109 | 4 |
| Troponin T Quantitative | < 13 | 2 |
| | [13, 36) | 0 |
| | ≥ 36 | 4 |
| Blood urea nitrogen (BUN) | < 4.7 | 0 |
| | [4.7, 8) | 2 |
| | [8, 16.4) | 4 |
| | ≥ 16.4 | 6 |
| Systolic BP | < 105 | 6 |
| | [105, 121) | 5 |
| | [121, 148) | 4 |
| | [148, 174) | 2 |
| | ≥ 174 | 0 |
| Sodium | < 95 | 8 |
| | [95, 100) | 5 |
| | [100, 104) | 3 |
| | [104, 107) | 0 |
| | ≥ 107 | 1 |
| Bicarbonate | < 18.8 | 3 |
| | [18.8, 27.2) | 0 |
| | ≥ 27.2 | 2 |
| Chloride | <95 | 8 |
| | [95, 100) | 5 |
| | [100, 104) | 3 |
| | [104, 107) | 0 |
| | ≥ 107 | 1 |
| **BAS#** | < 0.02 | 2 |
| | [0.02, 0.03) | 1 |
| | [0.03, 0.05) | 1 |
| | [0.05, 0.07) | 0 |
| | ≥ 0.07 | 1 |
| **RDW** | < 12.3 | 0 |
| | [12.3, 13.1) | 1 |
| | [13.1, 14.6) | 4 |
| | [14.6, 17.2) | 7 |
| | ≥ 17.2 | **10** |

**Figure S1.** Parsimony plot on the validation cohort based on AutoScore-Survival.

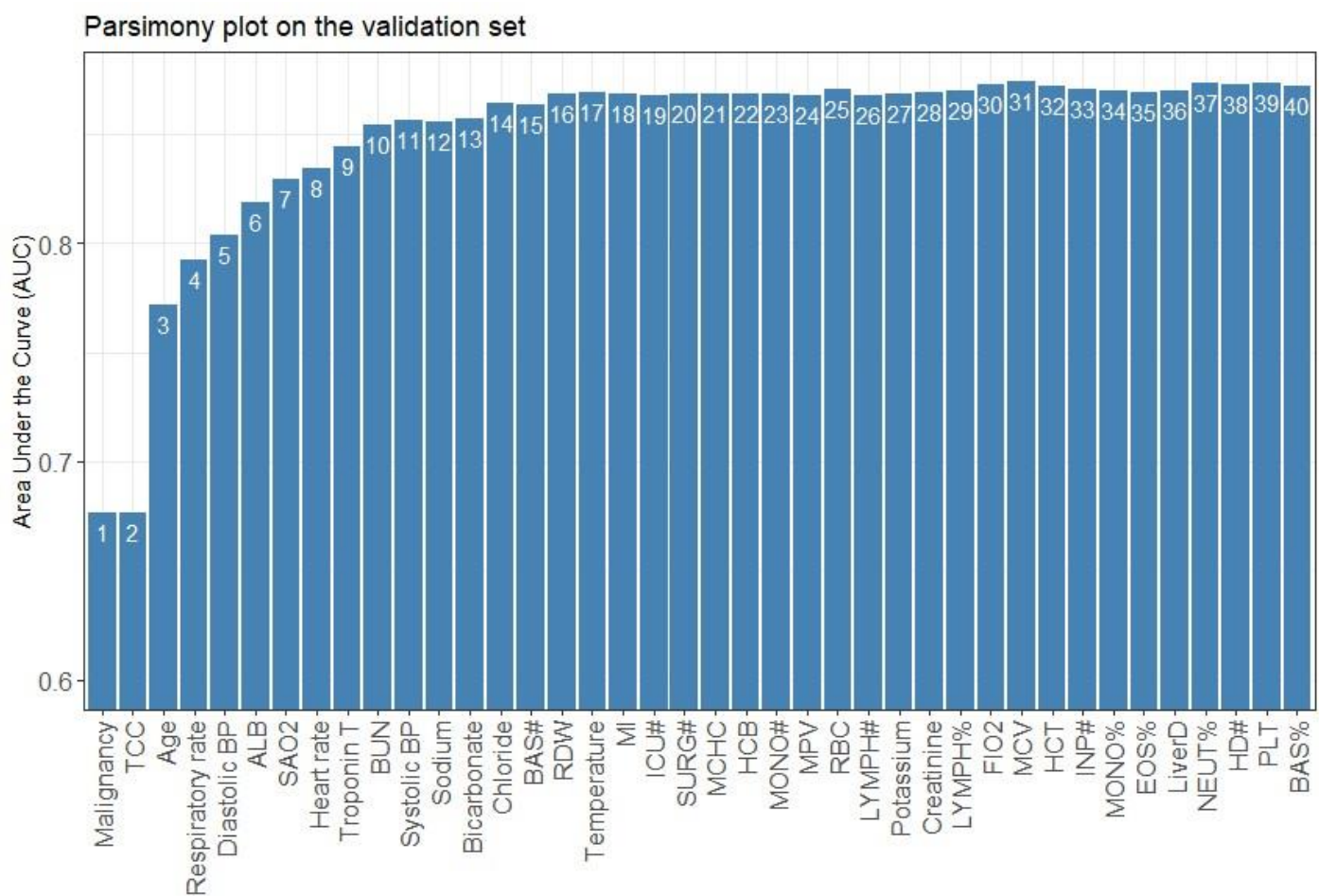

**Figure S2.** Choosing penalty strength $\alpha$ by concordance index (based on C-index).

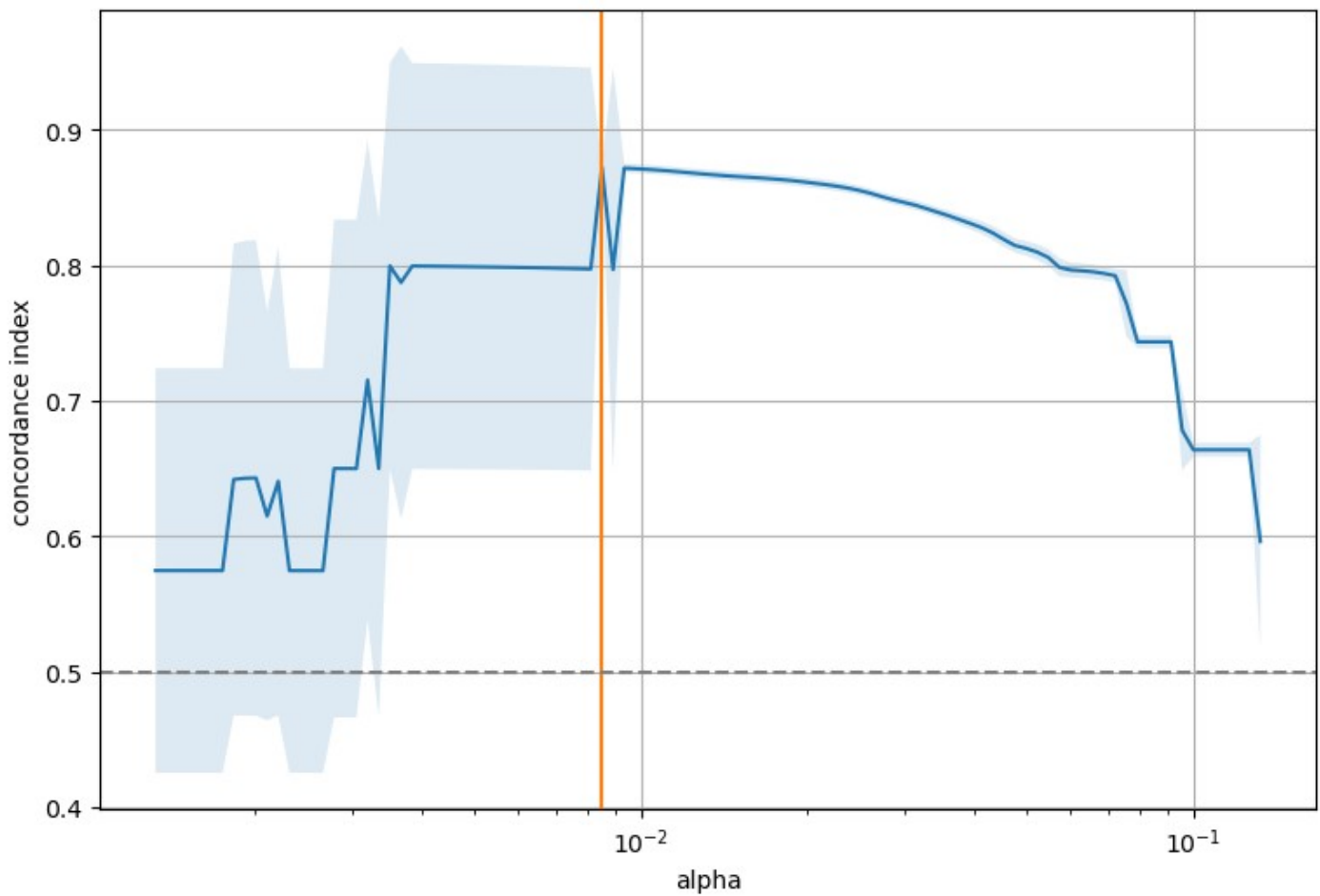

The figure shows that there is a range to the right for $\alpha$, where it becomes too large and sets all coefficients to zero. Conversely, if $\alpha$ becomes too small, an excessive number of features enter the model, leading to performance approaching that of a random model again. The optimal point, represented by the orange line, lies somewhere in the middle.

**Figure S3.** Variable importance on the validation cohort based on the CoxEN model.

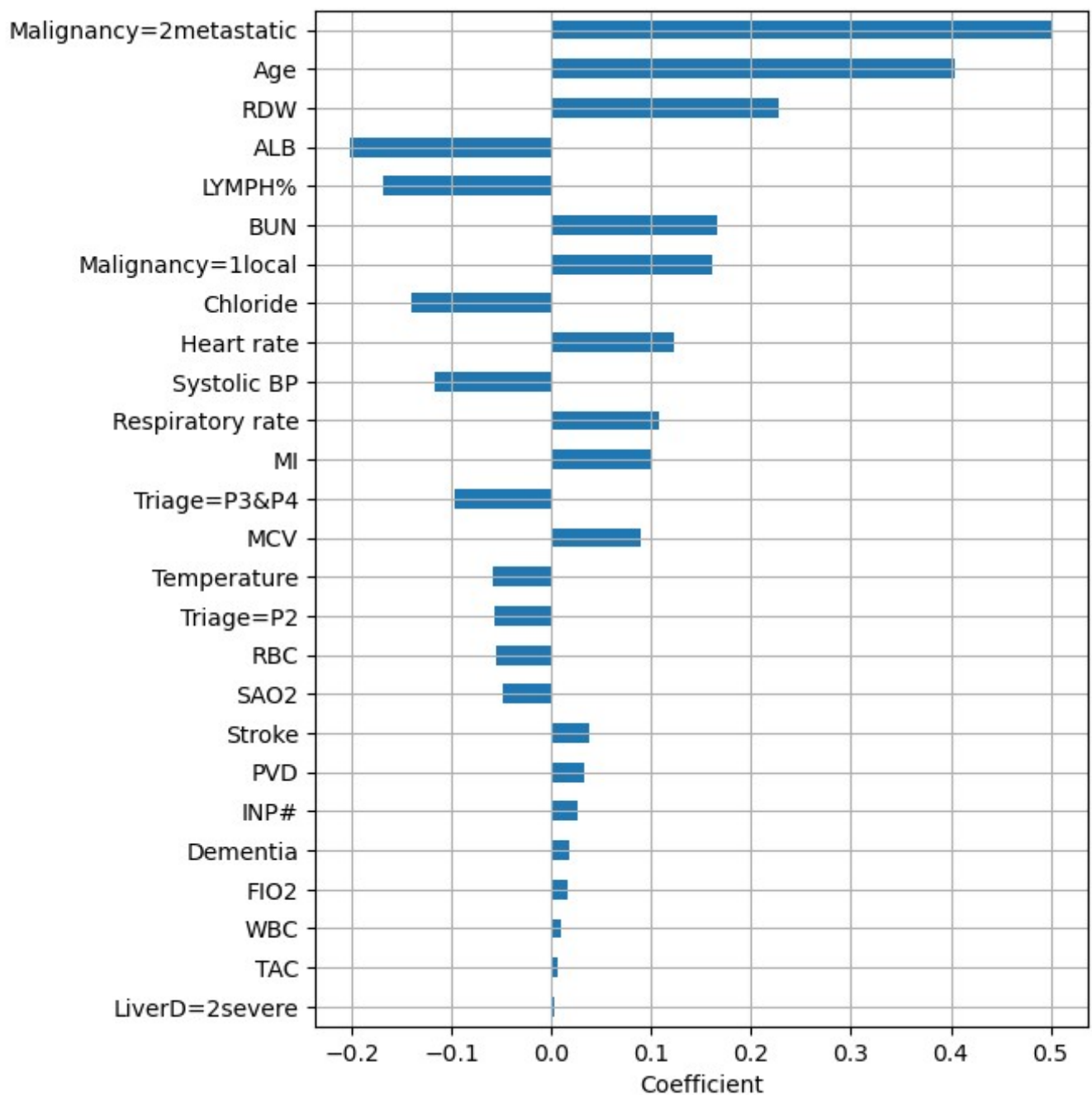

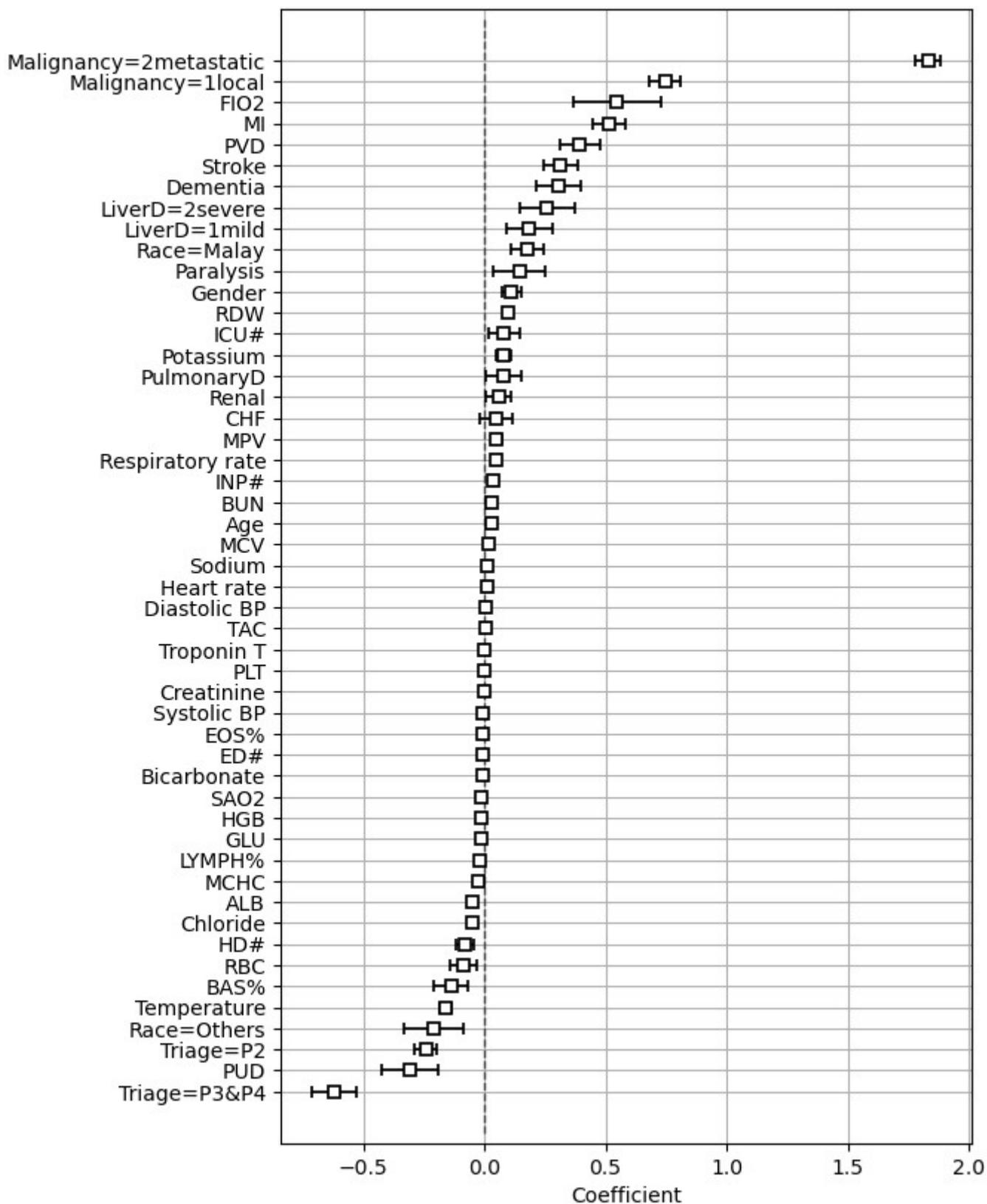

**Figure S4.** Variable importance based on the stepwise CoxPH method.

**Figure S5.** Variable importance based on RSF.

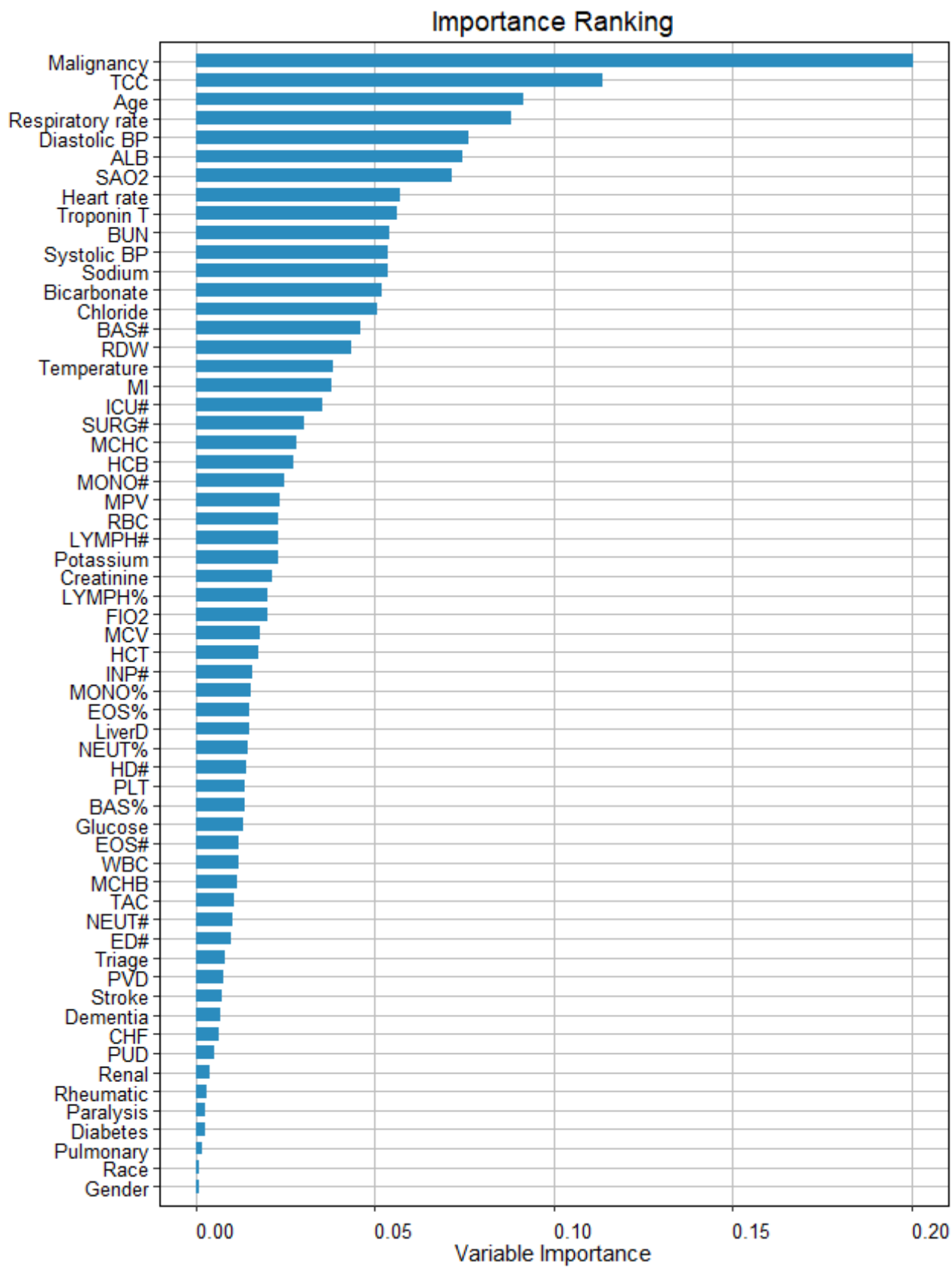

**Figure S6.** Variable importance based on GBM.

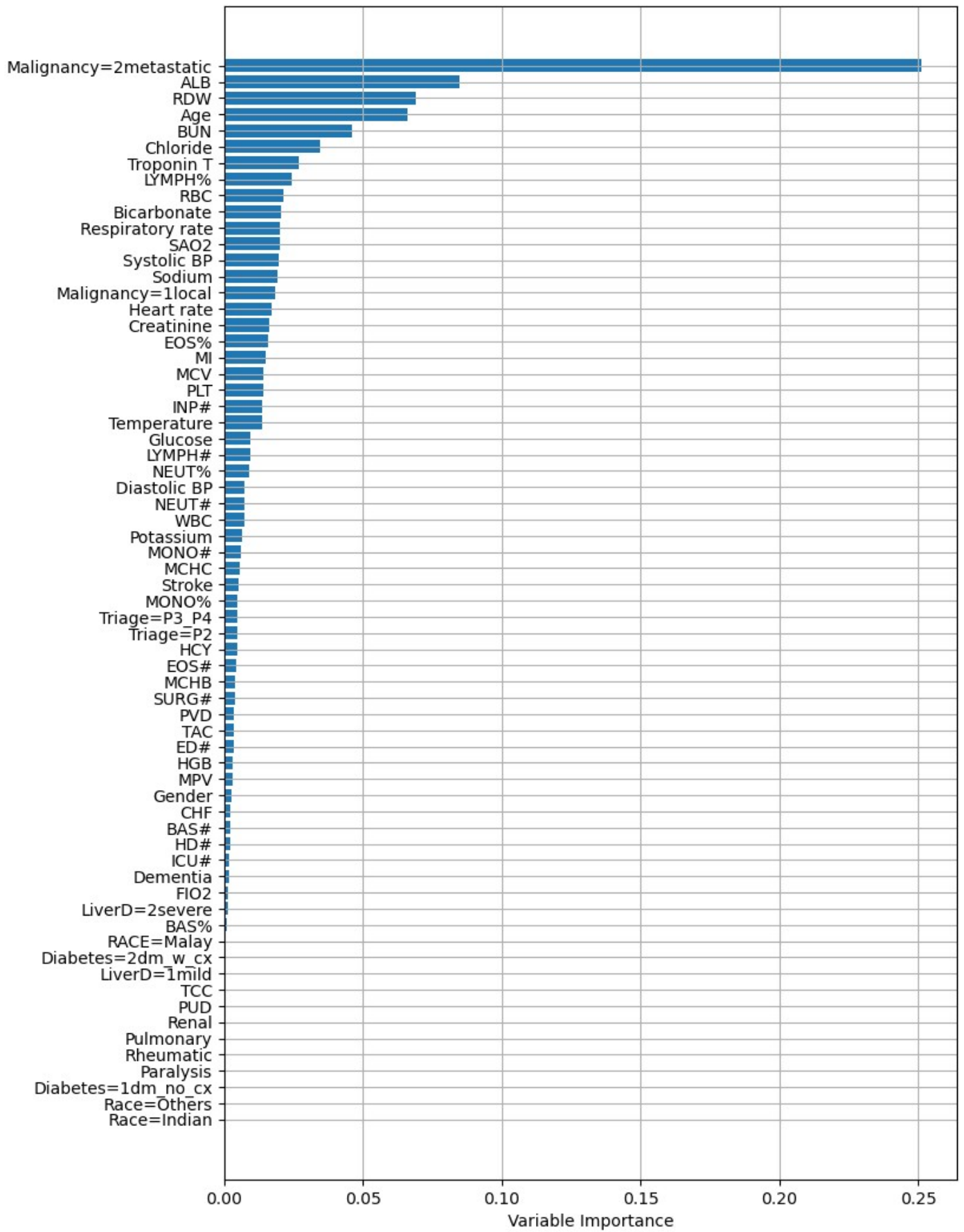

**Figure S7.** A plot about the loss of partial log-likelihood on training and validation cohort for DeepSurv algorithm.

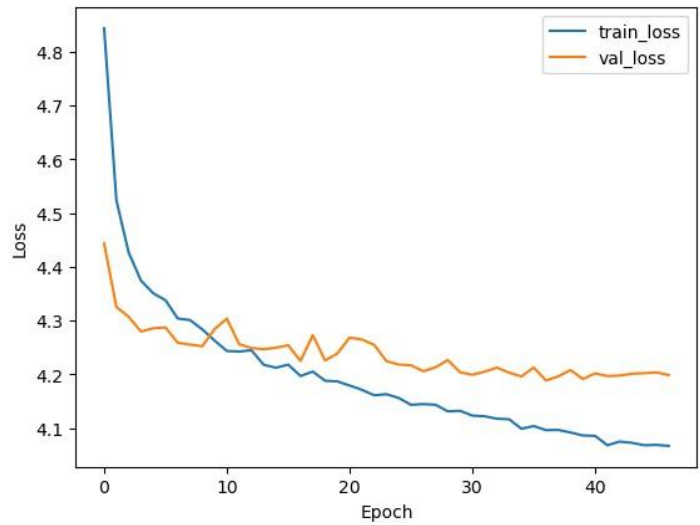

**Figure S8.** A plot about the loss of partial log-likelihood on training and validation cohort for CoxTime algorithm.

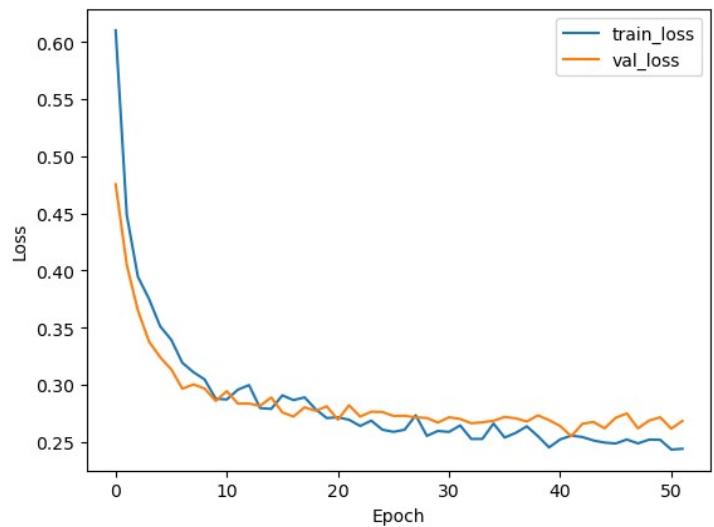

**Figure S9.** A plot about the loss of partial log-likelihood on training and validation cohort for DeepHit algorithm.

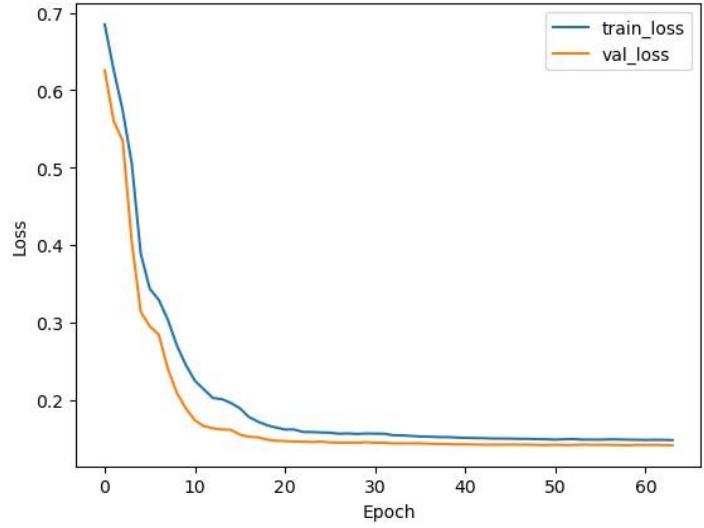

**S-Method**

Let R denote the risk score, without loss of generality, suppose that larger values of the risk score R are associated with greater hazards. The C-index can be defined as the proportion of concordance pairs among the population, $P(R_i > R_j | T_i < T_j)$, where $(R_i, T_i)$ and $(R_j, T_j)$ indicate two independent observations from two randomly chosen subjects $i$ and $j$. The empirical C-index can be defined as

$$\hat{C} = \frac{\sum_{i<j} \delta_i I(Y_i < Y_j) I(\hat{R}_i > \hat{R}_j) + \delta_j I(Y_j < Y_i) I(\hat{R}_j > \hat{R}_i)}{\sum_{i<j} \delta_i I(Y_i < Y_j) + \delta_j I(Y_j < Y_i)}.$$

The BS for survival data for a cohort I is define as

$$\text{BS}(t, \hat{S}) = N^{-1} \sum_{i \in I} \frac{1 - I(Y_j < t, \delta_i = 0)}{\hat{G}(Y_i \wedge t)} \{I(Y_j > t) - \hat{S}(t|X_i)\}^2,$$

where $\hat{G}(\cdot)$ is the Kaplan-Meier estimator of the survival function for the censoring times, and $I(\cdot)$ denotes an indication function. Then the IBS can be obtained by calculating the average of BSs over all observed time points $(0, \tau)$, which is given by $\text{IBS} = \tau^{-1} \int_0^\tau BS(u, \hat{S}) du$.